## 9.9. APPENDIX 9. Conference „Informacinės technologijos 2006"preceedings

# ENGLISH-LITHUANIAN-ENGLISH MACHINE TRANSLATION LEXICON AND ENGINE: CURRENT STATE AND FUTURE WORK

**Gintaras Barisevičius, Bronius Tamulynas**

*Kaunas University of Technology*

This article overviews the current state of the English-Lithuanian-English machine translation system. The first part of the article describes the problems that system poses today and what actions will be taken to solve them in the future. The second part of the article tackles the main issue of the translation process. Article briefly overviews the word sense disambiguation for MT technique using Google.

**1 Introduction**

The English-Lithuanian-English (ELE) dictionary-lexicon was chosen to be open to the user, so that he could modify the database on-demand. This dictionary contains Lithuanian and English words related to each other according to their meaning. However, this is not an ordinary dictionary and compare to it such dictionary contains much more attributes and morphological information of speech parts that are required for the MT. Currently, the lexicon supports all parts of speech for Lithuanian and English languages. Since the Lithuanian and English parts are strictly separated, so it is possible to extend the database by adding additional languages either for Lithuanian or English language [3].

Polysemy problem is also solved in the dictionary by adding an additional table between two tables linking different translations of the word in the target language. The translations for the same words are enumerated in descending priority in both directions. In this way it is possible to ensure that even if the translation won't be very exact, the user will be able to choose the suitable words himself and the system will choose the word with highest priority. Additionally, there is a possibility to include domain attributes for the nouns in the dictionary. That allows choosing the word during the translation not only by its translation priority but also by the domain, i.e. the word with the top domain goes the first [3].

The word entry is quite simple, since the lexicographer can see all generated morphological forms in the tables, which layout is so that it would be easy to check the correctness of all forms. That eases the lexicographer work and speeds up the word entrance process.

The Lithuanian Government approved to support this project according to the national program "Lithuanian language in Information society for the years 2005-2006 for the development of the Lithuanian language technologies including computer-based translation".

The first phase of this project for the year 2005 has been completed and the prototype of the lexicon has been finally created.

**2 Current system state**

**Lexicon.** Currently the database of the dictionary-lexicon contains 57 tables that already contain 250 words from Lithuanian to English and vice versa. Of course, each word requires much more than one record, since every morphological form is stored as a related record. So since Lithuanian noun has at least 14 forms and verb have more than 300 forms [4].

There is 11 part of speech windows for Lithuanian and 12 (one additional for auxiliary words and determiners) for English. For the moment the word can be entered to the dictionary only with its translation. When the phrase dictionary implemented, we will consider splitting the interface into two windows or panels which can be created for both languages or alone for one language.

The manual testing of the system has been performed for several months. Graphical user interface was tested by independent tester. A lot of errors were discovered and had to be fixed. Total sum of tested words reaches about 1000 words, but if you consider, that each word has at least several morphological forms and at least several attributes to be tested (here interjections, conjunctions and similar words which are not variable and not inflectional are excluded).

**Phrase dictionary.** Phrase dictionary was separated from the core word dictionary. The reason for that was the large number of phrases and if they were related physically to the word dictionary, when the word deleted all related phrases would be deleted as well. That is not very efficient, especially if the word deletion occurs accidentally. Besides, the phrases are stored into the same dictionary if you look from database management system view, but the tables do not relate to the lexicon tables. The phrase dictionary is still in implementation state, but the architectural decisions were already made.

**Data entry.** Theoretically data entry to the database could be performed parallel on-line by several lexicographers, but then there is "who did what" problem. If one of lexicographers involves the error into the database it is almost impossible to define which one is responsible for it. Of course, we could incorporate logging of each database modification according to the logged in user, but then we would get a huge overhead, since the same data should be replicated twice. Even, if we save only the reference to the modified data not replicating the data one more time, still we will have to access the database to save that reference. Due to that problem we decided that for the time being the data entry will be made locally by one lexicographer and the data later will be transferred to the on-line database.

The translation detects the words that are not in the dictionary, so monolingual text corpora will be possible to use, for extracting the words that are not in the dictionary. Then the words will be automatically passed to the chosen part of speech window for entrance. This is applicable either for Lithuanian or English language. This method of word entrance should be quite effective, since it is possible to choose the texts that contain the most frequent words in the language so that they would be entered to the database.

**Translation engine.** Current translation core uses direct translation and simple ending tuning according to grammar rules. Syntax rules are already incorporated into translation and they let to define which grammar structures are not allowed and which should be eliminated from the translation variants. However the transformational syntax rules are still being incorporated into the translation process.

Negations are not taken into account yet, but will be also incorporated into translation during the further development and improvement of translation engine.

The present state of translation engine would be not much in use for the real user, translating the texts neither from English nor Lithuanian, because the ending tuning is not entirely complete and word sense disambiguation is not incorporated yet. Besides, the tenses are not treated entirely correctly from Lithuanian language as well as from English. After some improvements first evaluations by independent tester could be already performed.

**3 Current problems and future work**

**Java Caching System.** Currently the connection to the database is straight-forward and doesn't use any additional caching, except for standard MySql cache. For the moment it is enough, since the system is in the testing state and doesn't require huge amounts of data to be processed, so the current MySql cache is enough. However, when the text size is large enough and if the database is in the remote host the caching on client side is needed, since the retrieval time from the local cache is shorter. For that purpose Java Caching System (JCS) can be used. As it is stated in [1] JCS is most useful for high read, but low put applications as it is exactly our system. And usage of the JCS noticeably decreases the latency time and the database is not a bottleneck in the system anymore [1]. The settings of the MySql database can be viewed by executing the following query (Table 13):

"SHOW VARIABLES LIKE '%query_cache%';"

**Table 13 MySql cache settings**

| Variable name | Value |
|---|---|
| have_query_cache | YES |
| query_cache_limit | 1048576 |
| query_cache_size | 26214400 |

Optimization. Optimization problem is always an issue when implementing a large system and should always be taken into account. If leaving out optimization, the system may result in long latency and unacceptable response time. During the automated testing very huge optimization problem was found. The table representing the list of words was working with an object array, which had to be recreated every time when new record added. When the number of words reached several thousands the insertion of each record took a large amount of time and it was clearly unacceptable. Very easy solution was made. The object array was replaced with an ArrayList, which obviously is faster than object array, especially when the new objects are consequently added to the list.

Another outstanding optimization problem is that when the word is looked up, all its morphological forms are return together with an object. Here the solution should as simple as implementing the additional queries returning only the required form of an object according to the word id number.

Semantics: Word Sense Disambiguation. "Word sense disambiguation is essential for the proper translation of words" as it is stated in [5]. Word sense disambiguation (WSD) process usually contains two steps that are: (1) determining all different senses for that word and (2) assigning the occurrences of a word to the appropriate sense [5].

Usually Word Sense disambiguation is performed manually, but this process is tedious and time consuming and today there are a number of techniques handling WSD, but most of them have those two steps mentioned above [5]. The second step requires information about the context of the word which is disambiguated and external knowledge sources [5], i.e. monolingual dictionary, encyclopaedia and etc.

In our MT system we have chosen using slightly different approach. That was done for two reasons. The latter information source is problematic to get, since there is not much encyclopaedias and monolingual dictionaries available in public that can be used and such disambiguation requires a lot computational power. As external knowledge source we will use monolingual text corpus, which can be quite effective performing word sense disambiguation for machine translation [7]. However, even monolingual text corpora for both English and Lithuanian are hard to get, even if they exist, but their usage is usually restricted only for research purposes. In addition to that, the different monolingual corpora usually tend to have different structures and we don't want to implement disambiguation algorithm for two different corpora. Here comes Google as a largest text database in the world, which has quite fast look-up and result display. Most importantly, Google displays the result number for each requested query. As it is stated in [6] Google can be used to find contextually relevant terms and their usage context.

In out MT system, actually we don't have to look up for different word senses (skipping step 1), since the translation gives the different senses for the word automatically as they are stored in the dictionary. So all we need only to choose the appropriate sense and as we mentioned before we are going to use Google for that purpose. There is an automated API for Google queries, but unfortunately it is limited to 1,000 queries per day and may return only 10 results per query. The total count fortunately is acquired this way. However 1,000 queries are not enough. For the beginning that should be enough to see the effectiveness of the algorithm and later if the usage of Google will be reasonable we will use indirect Google queries (not using API, but URL for queries) or we will have to extract our own monolingual corpora. We will have to decide which sense is most appropriate by calculating maximum likelihood estimation for the word sense with related words to it. For example, if we translate the sentence"pen is on the table" and then will look up all the senses in the Google (2 table) we will end up with such results (assuming that table has three meanings, and pen has also three meanings):

**Table 14 Possible sentence "pen is on the table" translations**

| Translation | Results by Google |
|---|---|
| Gulbė yra ant lentelės | 13 |
| Rašiklis yra ant lentelės | 16 |
| Areštinė yra ant lentelės | 5 |
| Gulbė yra ant stalo | 219 |
| Rašiklis yra ant stalo | 301 |
| Areštinė yra ant stalo | 18 |
| Gulbė yra ant plokščiakalnio | 0 |
| Rašiklis yra ant plokščiakalnio | 0 |
| Areštinė yra ant plokščiakalnio | 0 |

It is obvious from the results, that the correct translation is the fifth one. Of course, the fourth one is quite close, but considering that "Pen" sense as "Gulbė" is not likely to be used in technical texts so it won't be in our dictionary.

## 4 Conclusions

The lexicon and translation subsystems states were discussed in the article. It is obvious that the biggest current task is to collect large word dictionary. Next, we have to implement and also collect phrase dictionary. Phrase dictionary implementation will be performed parallel to translation engine implementation. The translation engine is only in its early stage and much work must be done there. Negations, tuning and sense disambiguation problems must be handled as well as syntax rule incorporation for transformation of the sentences must be finished implementing. When the phrase dictionary will be complete it will have to be incorporated into translation as well.

The data entry enhancement using text corpus was discussed and word sense disambiguation solution was briefly overviewed in the end of the article.